\documentclass[sigconf,nonacm,natbib=false]{acmart}
\renewcommand\footnotetextcopyrightpermission[1]{}

\AtBeginDocument{%
  }

\ifdefined\LaTeXML
  \usepackage[numbers]{natbib}
  \csname bibliographystyle\endcsname{plainnat}
  \setcitestyle{numbers,sort&compress}
  \newcommand{\printbibliography}{\csname bibliography\endcsname{ref}}
  \newcommand{\addbibresource}[1]{}
\else
  \RequirePackage[
    datamodel=acmdatamodel,
    style=acmnumeric,
    ]{biblatex}

  \DeclareSourcemap{
    \maps[datatype=bibtex]{
      \map{
        \step[fieldsource=entrykey, match=noauthor, final]
        \step[fieldset=presort, fieldvalue={zz}]
        \step[fieldset=options, fieldvalue={useauthor=false, useeditor=false}]
      }
    }
  }

  \renewbibmacro*{date}{\iffieldundef{year}{}{\printtext[parens]{\printdate}}}

\fi

\begin{document}

%% title
\title[Puda: Private User Dataset Agent for User-Sovereign and Privacy-Preserving Personalized AI]{Puda: Private User Dataset Agent \\ for User-Sovereign and Privacy-Preserving Personalized AI}

\author{Akinori Maeda}
\affiliation{
  \institution{Research Institute of Advanced Technology, SoftBank Corp.}
  \country{Japan}
}
\email{akinori.maeda@g.softbank.co.jp}

\author{Yuto Sekiya}
\affiliation{
  \institution{Research Institute of Advanced Technology, SoftBank Corp.}
  \country{Japan}
}
\email{yuto.sekiya@g.softbank.co.jp}

\author{Sota Sugimura}
\affiliation{
  \institution{Research Institute of Advanced Technology, SoftBank Corp.}
  \country{Japan}
}
\email{sota.sugimura01@g.softbank.co.jp}

\author{Tomoya Asai}
\affiliation{
  \institution{WebDINO Japan}
  \country{Japan}
}
\email{dynamis@dynamis.jp}

\author{Yu Tsuda}
\affiliation{
  \institution{Turnt Up Technologies, Inc.}
  \country{Japan}
}
\email{tsuda@turntup.co.jp}

\author{Kohei Ikeda}
\affiliation{
  \institution{TechArts Co., Ltd.}
  \country{Japan}
}
\email{godai@techarts.co.jp}

\author{Hiroshi Fujii}
\affiliation{
  \institution{Acutus Software, Inc.}
  \country{Japan}
}
\email{fujii@acutus.jp}

\author{Kohei Watanabe}
\affiliation{
  \institution{WebDINO Japan}
  \country{Japan}
}
\email{watanabe@webdino.org}

\renewcommand{\shortauthors}{Maeda et al.}

%% article.
\begin{abstract}
  Personal data centralization among dominant platform providers---including search engines, social networking services, and e-com\-merce---has created siloed ecosystems that restrict user sovereignty, thereby impeding data use across services. Meanwhile, the rapid proliferation of Large Language Model (LLM)-based agents has intensified demand for highly personalized services that require the dynamic provision of diverse personal data. This presents a significant challenge: balancing the utilization of such data with privacy protection. To address this challenge, we propose Puda (Private User Dataset Agent), a user-sovereign architecture that aggregates data across services and enables client-side management. Puda allows users to control data sharing at three privacy levels: (i) Detailed Browsing History, (ii) Extracted Keywords, and (iii) Predefined Category Subsets. We implemented Puda as a browser-based system that serves as a common platform across diverse services and evaluated it through a personalized travel planning task. Our results show that providing Predefined Category Subsets achieves 97.2\% of the personalization performance (evaluated via an LLM-as-a-Judge framework across three criteria) obtained when sharing Detailed Browsing History. These findings demonstrate that Puda enables effective multi-granularity management, offering practical choices to mitigate the privacy--personalization trade-off. Overall, Puda provides an AI-native foundation for user sovereignty, empowering users to safely leverage the full potential of personalized AI.
\end{abstract}

\maketitle

\section{Introduction}
In online markets such as search, social networking services (SNS), and e-commerce, personal data is often enclosed within dominant platform operators, resulting in a siloed structure that hinders cross-service utilization \cite{kranz_data_2023, kuebler-wachendorff_right_2021}. Under this structure, governance frequently relies on notice-and-control mechanisms based on terms of service; however, it is difficult for users to continuously manage their data on this basis while understanding complex data flows \cite{solove_privacy_2013}, making user-sovereign personal data management hard to achieve.

Conversely, the advent of Large Language Model (LLM)-based agents has surged the demand for diverse personal data to fuel highly personalized services. The rapid evolution of LLMs, such as ChatGPT \cite{openai_gpt-4_2024} and Gemini \cite{comanici_gemini_2025}, has transcended simple task resolution, giving rise to autonomous AI agents capable of executing complex workflows. Furthermore, interconnection protocols like Agent2Agent (A2A) \cite{noauthor_agent2agent_2025}, Model Context Protocol (MCP) \cite{noauthor_model_2024}, and Agent Communication Protocol (ACP) \cite{noauthor_agent_2025} are accelerating the trend toward comprehensive solutions where agents collaborate across diverse services. In this agent-centric ecosystem, accessing a user's rich context—spanning information across multiple services—is prerequisite for agents to accurately understand user preferences and intent. However, providing personal data for personalization inevitably invokes the trade-off between utility and privacy protection \cite{herlocker_evaluating_2004, zhu_privacy_2017}. The necessity and sensitivity of data shared with AI agents are not static; they fluctuate dynamically depending on the task and context. Consequently, uniform management approaches like notice-and-control are insufficient for the flexible demands of modern AI agents. To address this, recent studies have proposed frameworks that delegate privacy protection to agents \cite{zhang_towards_2025}, acting as proxies that learn user privacy literacy to control data disclosure. While promising for their autonomous flexibility, these approaches suffer from a fundamental limitation: reliance on the probabilistic nature of LLMs introduces uncertainty, making it impossible to deterministically prevent the unintended leakage of sensitive information. Furthermore, existing literature lacks a discussion on how to technically collect and manage the diverse, cross-service personal data required for such personalization under a user-sovereign framework. There is an urgent need for a data provision method that effectively balances personalization and privacy while ensuring user sovereignty.

To bridge this gap, we propose "Puda" (Private User Dataset Agent), a user-sovereign, multi-granularity personal data management architecture. Puda collects and manages personal data across services within a user-controlled environment—specifically, a browser-based platform—and organizes this data into three granular privacy levels: Detailed Browsing History, Extracted Keywords, and Predefined Category Subsets. While leveraging an LLM in the preprocessing stage to enable flexible content extraction, the extraction of categories at the highest privacy level is designed to restrict the output space deterministically by limiting it to a subset of predefined categories. A distinct feature of Puda is that it empowers users to select the granularity of data provided to external agents. This architecture resolves the probabilistic uncertainty inherent in autonomous privacy-preserving agents while offering effective options for users. 

We evaluated the efficacy of the personal data context provided by Puda using a travel planning task \cite{singh_personal_2024, xie_travelplanner_2024}, a common scenario for assessing complex reasoning and personalization in AI agents. Our evaluation clarifies the impact on personalization performance as well as practical costs such as latency and token consumption. The contributions of this study are threefold:
\begin{enumerate}
\item We propose an architecture that collects personal data via a browser-centric approach and manages personal data across services, providing data at multi-granular privacy levels rooted in user sovereignty (Section 3).
\item We implemented the Puda architecture assuming a personalized AI scenario for a travel planning agent task (Section 4).
\item Through empirical evaluation, we demonstrated that the multi-granularity data management provided by Puda is effective for personalized AI in terms of performance, privacy protection, and practical cost (Sections 5 and 6).
\end{enumerate}

\section{Related Work}
\subsection{User Sovereignty in Data Ecosystems}
Dominant platform operators aggregate personal data, creating silos that hinder cross-service utilization under a user-sovereign framework \cite{kranz_data_2023, kuebler-wachendorff_right_2021}. To address this, architectures known as Personal Data Stores (PDS) or Personal Information Management Systems (PIMS) \cite{european_data_protection_supervisor_personal_2020, janssen_personal_2020}—such as Solid \cite{european_data_protection_supervisor_personal_2020, mansour_demonstration_2016} and MyData \cite{poikola_mydataglobaldeclaration_2017}—have been proposed to decouple applications from data storage. While these initiatives provide foundational infrastructure for data portability, they have not yet addressed the critical challenge of effectively provisioning unstructured user context to personalized AI. 

\subsection{Privacy in Personalized AI}
  Since 2020, the task adaptation design for LLM-based agents has shifted from individual model training to inject user context and external knowledge during inference \cite{edemacu_privacy_2025}. The effectiveness of In-Context Learning (ICL) has been demonstrated alongside the scaling of large models \cite{brown_language_2020}. Similarly, Retrieval-Augmented Generation (RAG) \cite{lewis_retrieval-augmented_2020} has become widely adopted. Furthermore, to support long-term tasks and continuous dialogue, memory-oriented agents have been proposed \cite{packer_memgpt_2023, park_generative_2023}. While these trends enhance utility, they are also pointed out to increase privacy risks due to the utilization of diverse contexts \cite{wen_membership_2024, zeng_good_2024}. These risks are further exacerbated by standard interconnection protocols such as MCP, A2A, and ACP, which expand the scope of agent collaboration with external tools and other services \cite{habler_building_2025, hou_model_2025, kong_survey_2025, louck_security_2025}.

In response to the growing need for privacy measures in AI agents, researchers propose agent-mediated frameworks acting as proxies to mediate data provision. These include learning preferences to balance utility and privacy via Contextual Integrity \cite{nissenbaum_privacy_2004, zhang_towards_2025} generating privacy rules from user edits \cite{guo_privi_2025}, and updating consent iteratively upon detecting data provision events \cite{slate_iterative_2025}.

However, these agent-mediated privacy protection approaches face fundamental challenges. First, regarding probabilistic uncertainty, since judgments made by AI agents involve inherent uncertainty, misjudgments regarding privacy protection are unavoidable. Second, the cold start problem leads to potential erroneous inferences at the initial stage due to scarce data regarding user privacy preferences. Third, inconsistency in user behavior complicates protection, as studies report that even privacy-conscious users may inadvertently disclose sensitive information during interactions \cite{ngong_protecting_2025}. Consequently, blindly trusting learned privacy preferences risks reinforcing past suboptimal user judgments.

\subsection{Positioning of Puda}
In Puda, we address the user sovereignty discussed in Section 2.1 through an architecture in which users manage, on the user side, cross-service personal data centered on the browser. To tackle the challenges described in Section 2.2 regarding providing data for personalized AI, we attempt a solution by managing data under multiple privacy levels with different granularities, including a mechanism that deterministically constrains the output space.

\section{Puda Architecture}
\begin{figure}[t]
  \centering
  \includegraphics[width=\linewidth]{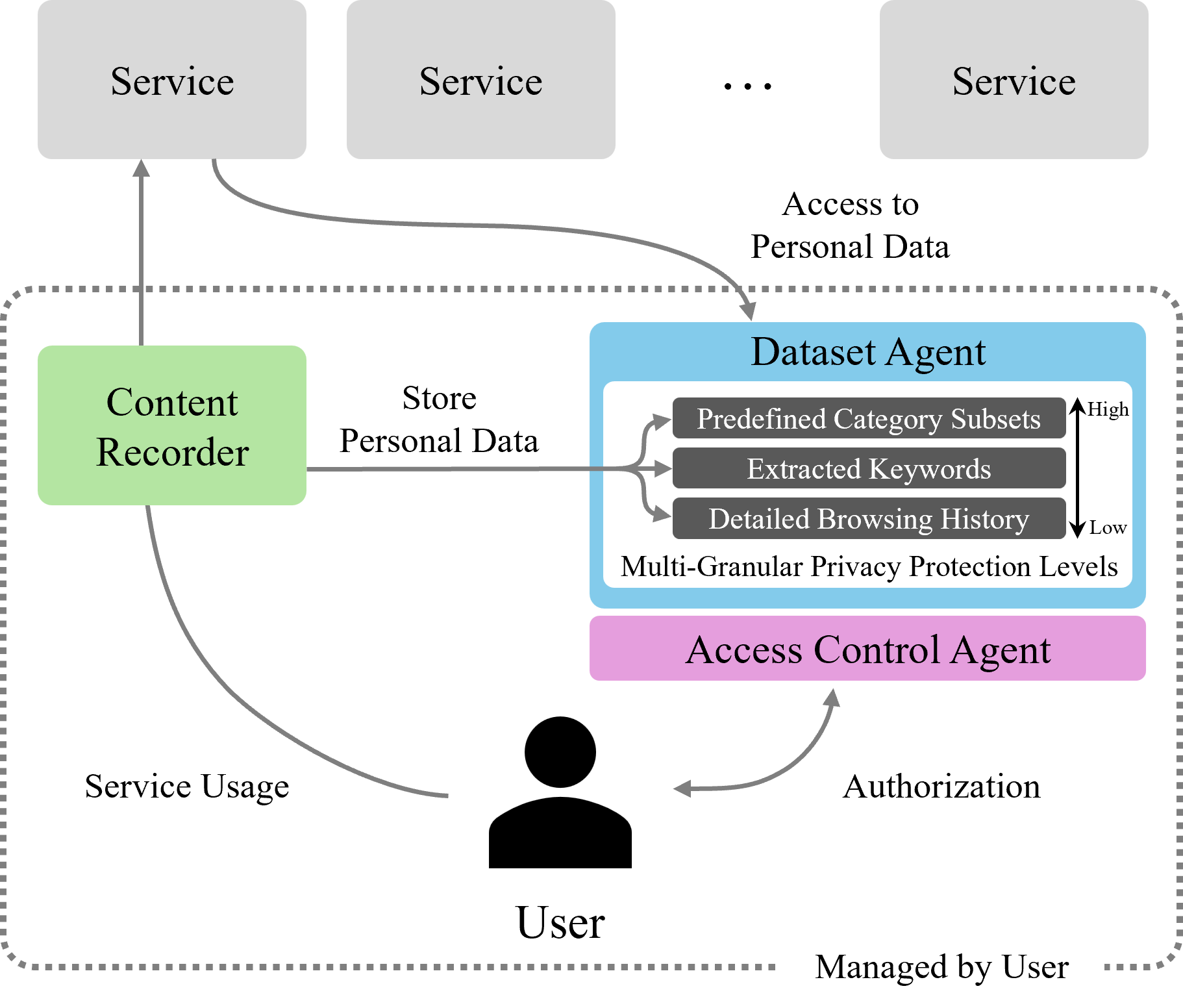}
  \caption{The Content Recorder collects users’ cross-service online activities. The Dataset Agent transforms this data into multi-granular datasets at different privacy levels. The Access Control Agent ensures that data shared with external services is scoped to the granularity authorized by the user.}
  \Description{}
\end{figure}
\subsection{Overview}
Figure 1 presents a schematic overview of the Puda architecture. This architecture comprises three primary components: the Content Recorder, the Dataset Agent, and the Access Control Agent. These components are designed to execute within a user-controlled environment, such as a local device, a private cloud, or Mobile Edge Computing (MEC) infrastructure. First, the Content Recorder captures personal data by recording user activities across multiple services. Subsequently, the Dataset Agent stores the acquired data, organizing it into multi-granular privacy levels. By managing data across distinct tiers— Predefined Category Subsets, Extracted Keywords, and Detailed Browsing History —the architecture ensures a design with low uncertainty, empowering users to explicitly select the desired granularity for disclosure. Finally, data provision to external services for personalization purposes is executed solely upon authorization mediated by the Access Control Agent. To ensure interoperability and scalability, communication between Puda components and collaboration with external agents adopt standard specifications, such as the A2A protocol.

\subsection{Browser-based Content Recorder}
The Content Recorder serves as the component responsible for recording user activities across services. It is implemented as a browser-based system. Given that the web browser functions as the pivotal interface for modern online services, this approach enables the comprehensive collection of interaction histories across diverse services through a single-entry point. Specifically, the Content Recorder captures the URL, page title, and the HTML body of visited pages. This data constitutes the raw logs, representing the highest level of sensitivity within the Puda design. All subsequent processing—specifically, the rigorous privacy protection measures required to generate multi-granular personal data from these raw logs—is exclusively delegated to the Dataset Agent. It should be noted that the current design of the Content Recorder focuses exclusively on browser-based activities; user behaviors occurring outside the browser environment, such as native mobile application usage or OS-level operations, remain outside the scope of this study.

\subsection{Dataset Agent Processing}
To enable granular privacy choices based on the user’s intent, the Dataset Agent transforms personal data into representations governed by multi-granular privacy levels. In this design, we adopt as a guiding principle that users should be able to select from a wide range of privacy protection levels and therefore define and manage the following three privacy protection levels.

\begin{itemize}
\item Detailed Browsing History: This comprises a list of URLs, titles, and summary data of pages visited by the user. Since the history data is visible in nearly its raw form, it entails a very high privacy risk; however, it is data that allows for a detailed grasp of user behavior.
\item Extracted Keywords: This comprises a set of keywords extracted from the user's browsing history. While identifying specific visited pages or services is difficult, this level poses a moderate privacy risk as sensitive proper nouns, such as specific location names or medical conditions, may still be extracted. However, this data facilitates the interpretation of abstract user preferences.
\item Predefined Category Subsets: This comprises a subset extracted from a predefined category list based on the user’s browsing history. Since the data is constructed solely from this list, the risk of leaking unexpected proper nouns is deterministically eliminated, offering the lowest privacy risk. Conversely, expressiveness is limited, as content outside the list cannot be represented.
\end{itemize}

Figure 2 illustrates the processing flow for generating the three types of data described above. The data processing workflow is broadly divided into two stages: per-page processing and per-user processing. In the per-page processing stage, an LLM generates a summary, and per-page keywords based on the HTML body of each page. Each keyword is assigned a sentiment label and a score, which are subsequently utilized by the LLM during inference. In the per-user processing stage, the data processed at the page level is aggregated into user-level data. The URLs, titles, and summaries are consolidated to form the Detailed Browsing Contents. Similarly, the keywords are aggregated and treated as user-level Extracted Keywords. Regarding the Predefined Category Subsets, an LLM uses the Detailed Browsing Contents and Extracted Keywords as input to extract items from a predefined category list that align with the user's preferences, thereby constructing a subset of the category list.
\begin{figure}[t]
  \centering
  \includegraphics[width=\linewidth]{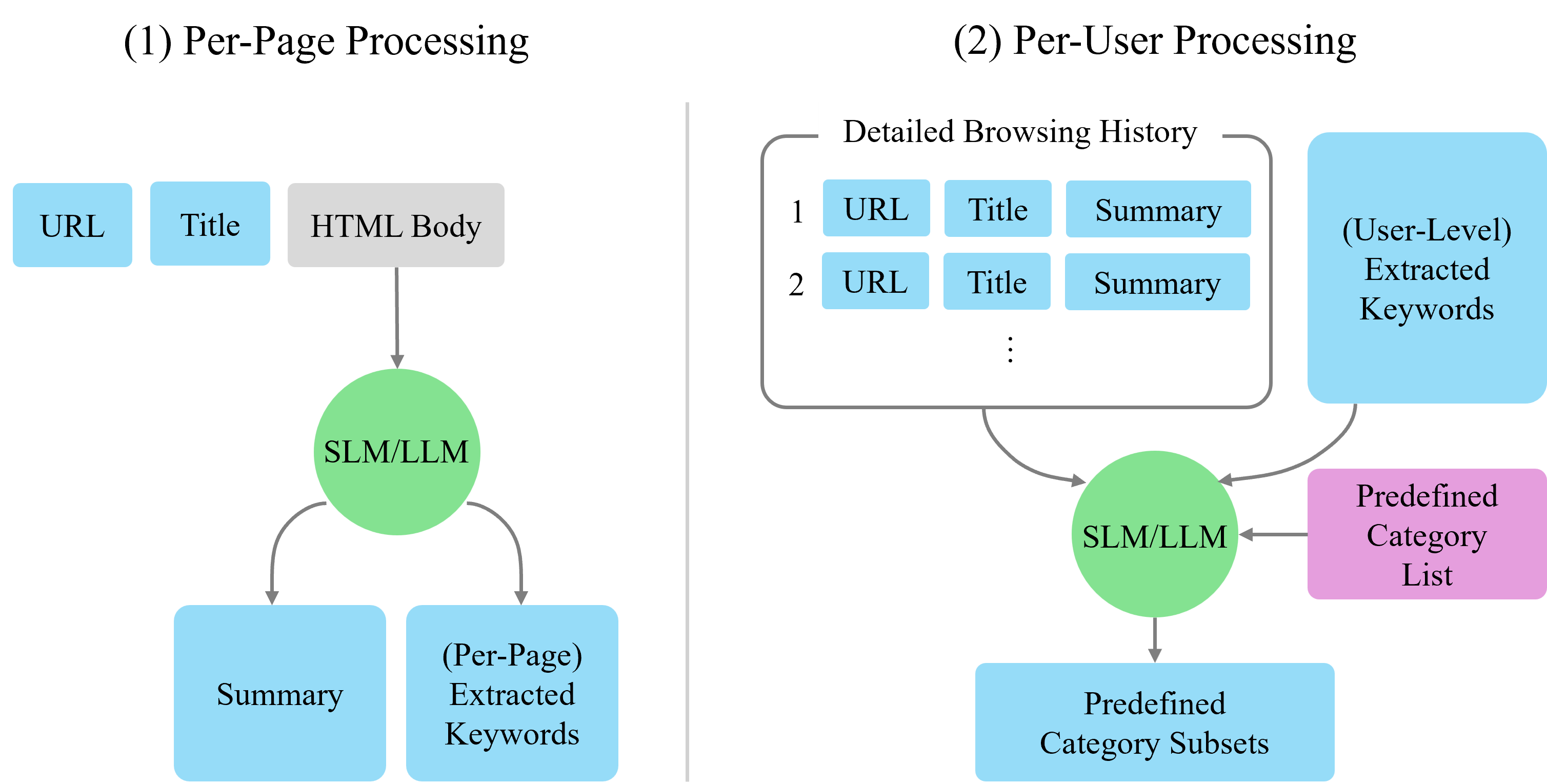}
  \caption{Data processing flow within the Dataset Agent transforming user history data into multi-granular privacy levels. The left side depicts the per-page processing, while the right side illustrates the per-user processing, which aggregates the page-level data into user-specific datasets.}
  \Description{}
\end{figure}

\subsection{Authorization Design}
In this architecture, data access control is designed using standardized specifications such as OAuth 2.0 and OpenID Connect. Crucially, the Dataset Agent (functioning as the Resource Server) and the Access Control Agent (functioning as the Authorization Server) may be operated by distinct administrative entities. Consequently, an authorization mechanism is required to establish trust even between these internal components.

Specifically, the design leverages OpenID Connect Discovery \cite{sakimura_openid_2023} and the Authorization Code Flow \cite{hardt_oauth_2012} to achieve this. The process initiates when the user, acting as the Resource Owner, provides the location of the Access Control Agent to the Dataset Agent. The Dataset Agent then retrieves the Access Control Agent's endpoints via OpenID Connect Discovery. Subsequently, the Authorization Code Flow is executed, and an access token is issued from the Access Control Agent to the Dataset Agent based on user authorization. Using this access token, the Dataset Agent registers information, such as its own endpoints, with the Access Control Agent. Following this registration, the user provides the location of the Access Control Agent to the external AI Agent. Through a similar procedure, the Access Control Agent issues an access token scoped for the Dataset Agent's endpoints and provides it to the AI Agent. By utilizing this access token, the AI Agent can retrieve personal data from the Dataset Agent strictly within the scope permitted by the user's authorization. In this study, to prioritize the discussion on personalization performance, the authorization mechanism is limited to the design phase, and its implementation is excluded from the scope of this work.
\begin{figure}[b]
  \centering
  \includegraphics[width=\linewidth]{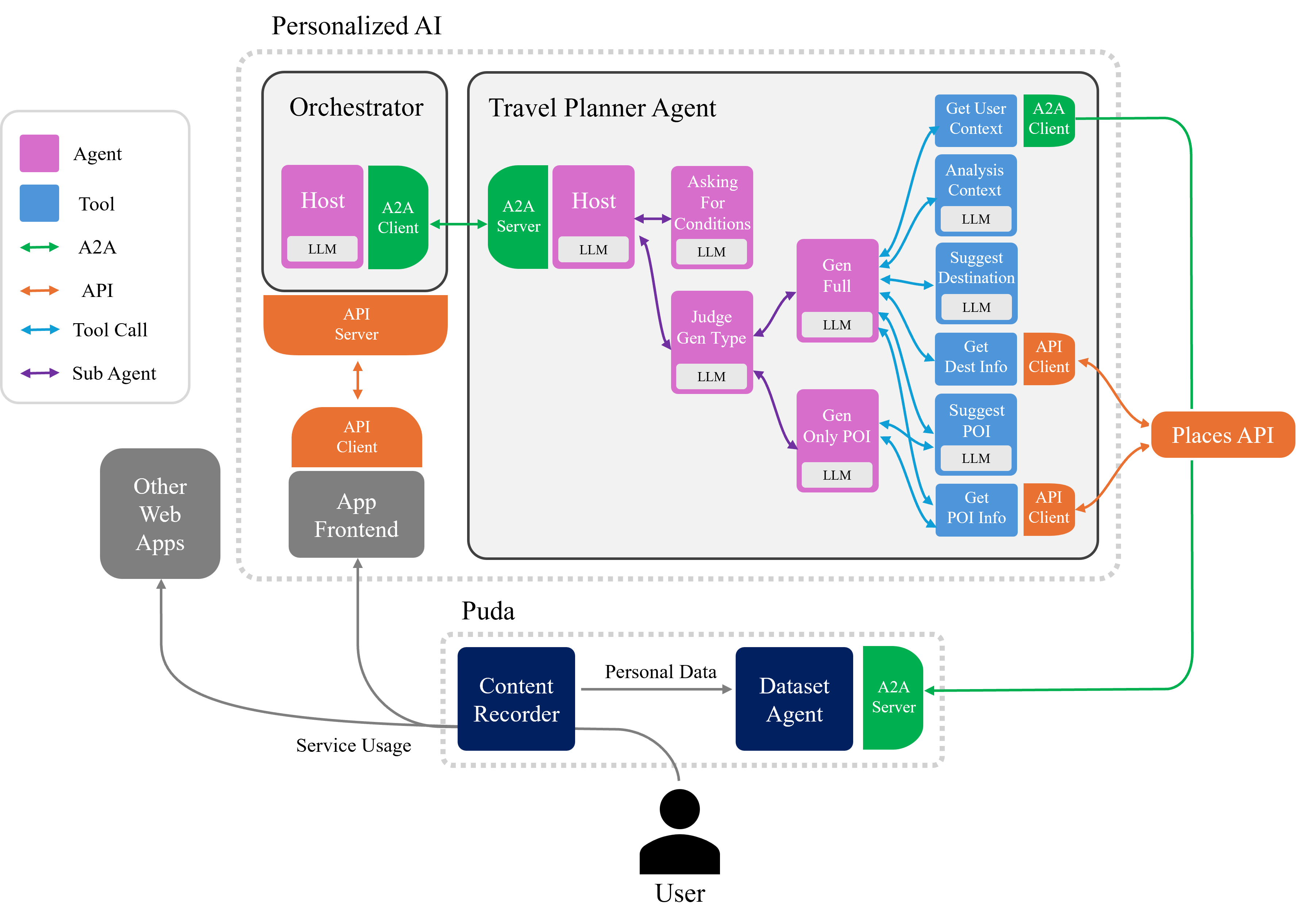}
  \caption{The User initiates a travel planning request via the App Frontend. The Orchestrator delegates this task to the Travel Planner Agent. The Travel Planner Agent executes the task by collaborating with internal Sub Agents and Tools. Personal data managed by Puda is provisioned to the Travel Planner Agent via the A2A protocol.}
  \Description{}
\end{figure}
\section{Implementation}
\subsection{Puda System}
The Content Recorder was implemented as a web browser extension designed to collect user interaction history. The collected data is transmitted to the server hosting the Dataset Agent, where it is processed into multi-granular privacy levels.

For the per-page processing, the extraction of summaries and per-page keywords was performed using Gemma 3 4B \cite{team_gemma_2025}. This model was selected as a Small Language Model (SLM) capable of being managed within a user-controlled environment. Regarding the summaries, we generated two variations: a long version, which utilizes the raw generation output, and a short version, produced by re-summarizing the content to achieve efficient context compression. For the generation of Predefined Category Subsets, we utilized GPT-5 nano \cite{openai_gpt-5_2025}. We acknowledge that employing a cloud-based model introduces a limitation regarding the strictly user-side processing ideal of the Puda architecture. However, this summarization task necessitates processing extensive user contexts, including Detailed Browsing History and Extracted Keywords, which exceeds the capabilities of many current on-device models. Considering the rapid trajectory of SLM advancements regarding context window expansion and reasoning efficiency \cite{microsoft_phi-4-mini_2025, team_gemma_2025}, we selected GPT-5 nano as a functional proxy to simulate future high-capability SLMs that will be deployable within user-controlled environments.

For the inputs used in Predefined Category Subsets creation, we utilized the long version of the summaries to provide rich information. Similarly, for user-level Extracted Keywords, no threshold filtering based on scores was applied; all data, accompanied by their respective scores and sentiment labels, were included in the input. The predefined category list was constructed based on the Content Categories provided by the Google Cloud Natural Language API \cite{google_cloud_content_2026}. Since the subjects of the evaluation experiment were Japanese, the categories were localized into Japanese. The hierarchy consists of 26 categories in the first tier, 256 in the second, and 810 in the third.

In Section 5, we evaluate the performance—alongside practical costs such as latency and token consumption—of various configurations, including the use of long/short summaries for Detailed Browsing History, Predefined Category Subsets creation using Extracted Keyword scores, and the generation of subsets with varying levels of abstraction by utilizing specific tiers of the predefined categories.
\subsection{Use Case: Travel Planner Agent}
Travel planning tasks are widely employed to evaluate personalization performance, as they require addressing multi-constraint planning scenarios that are heavily influenced by user preferences \cite{singh_personal_2024, xie_travelplanner_2024}. Following this precedent, we adopted a Travel Planner Agent as a use case for an external service and implemented a configuration where personal data is provisioned from the Dataset Agent.
\begin{table*}[t]
  \caption{For each condition of multi-granular personal data, we present the granularity level, the level of privacy protection, and a description of the data. Conditions in higher rows provide higher levels of privacy protection, whereas conditions in lower rows provide lower levels of privacy protection. In addition, within Categories, the level of privacy protection decreases in the order of Levels 1, 2, and 3; within Keywords, it decreases in the order of thresholds 0.90, 0.85, 0.80, and 0.75; and within Browsing History, it decreases in the order of Short and Long.}
  \label{tab:table1}
  \centering
  \includegraphics[width=\textwidth]{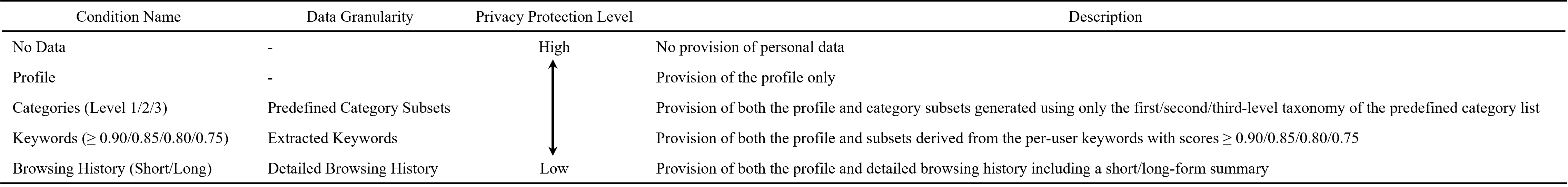}
\end{table*}

Figure 3 illustrates the system configuration of the Travel Planner Agent, including its integration with Puda. This agent was designed based on "travel-concierge" \cite{google_adk-samples_2025}, a sample agent from the Agent Developer Kit (ADK) \cite{google_agent_2025}. Anticipating future collaboration across diverse tasks beyond mere travel planning, the architecture incorporates an Orchestrator that mediates interactions between the User and the agents. The Travel Planner Agent accepts the travel budget and duration as input parameters. Each constituent agent operates using Gemini 2.5 Flash \cite{google_gemini_2025}, selected for its lightweight and practical performance. These task-specific agents collaborate via the A2A protocol to generate an output adhering to a strict JSON output schema. This schema comprises inferred user interests, three candidate destinations, three Points of Interest (POIs) per destination, and the rationale for these recommendations. Specifically, in addition to basic information such as location names, the output includes geographic coordinates (latitude and longitude) retrieved through integration with the Places API \cite{google_overview_2026}. Figure 4 presents the User Interface implemented for the Travel Planner Agent. All information displayed on the screen is rendered based on the JSON file generated by the Travel Planner Agent. We implemented this travel planning agent at a practical level, ensuring compatibility with application frontend. In Section 5, we clarify the trade-offs among personalization, privacy levels, and practical costs by varying the user context provided to this Travel Planner Agent according to multi-granular privacy levels.
\begin{figure}[b]
  \centering
  \includegraphics[width=\linewidth]{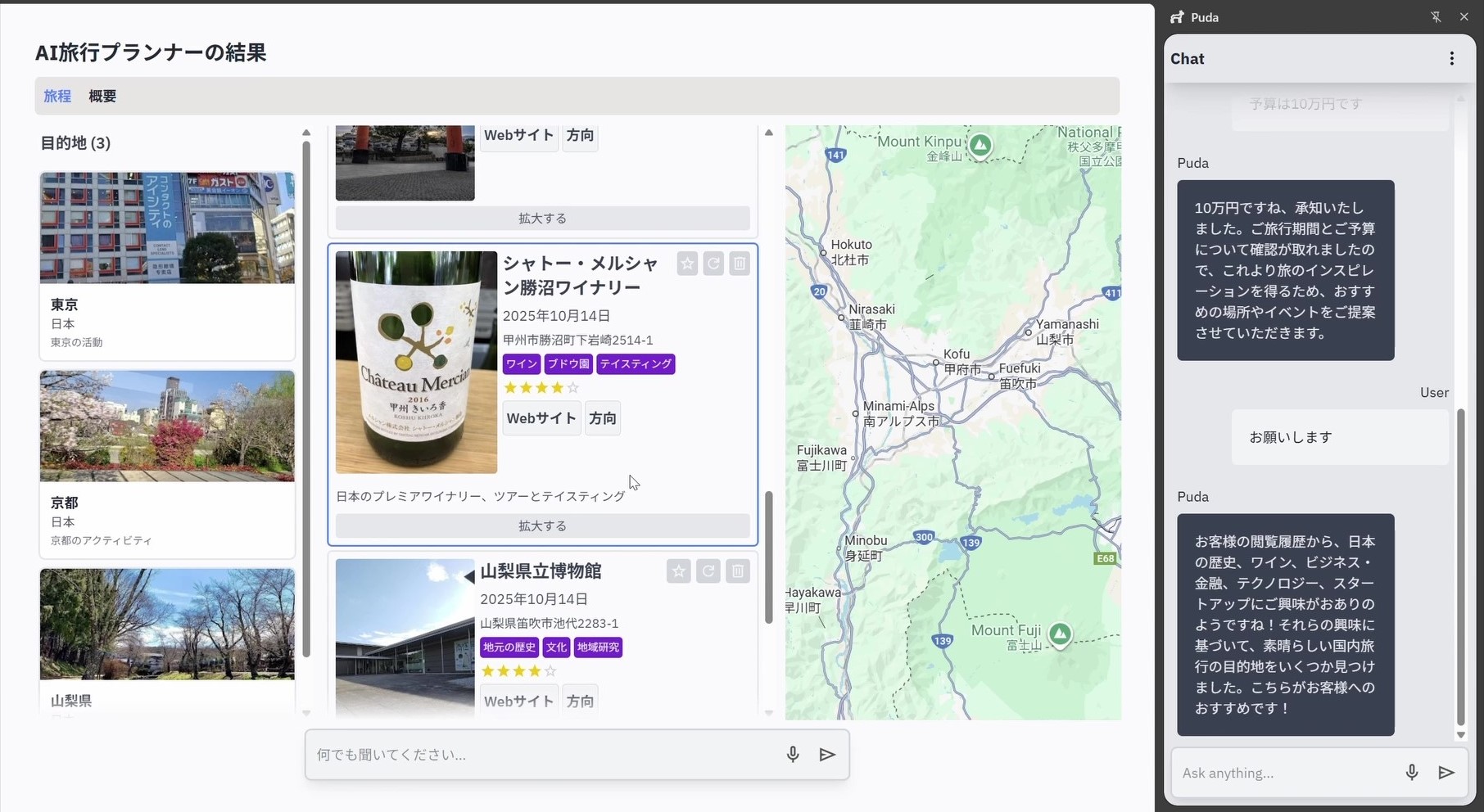}
  \caption{Japanese user interface of the Travel Planner Agent. Users request a travel plan via the chat interface on the right. The left panel displays three proposed candidate destinations, while Points of Interest (POIs) are presented in the center. In this figure, a winery in Yamanashi is highlighted.}
  \Description{}
\end{figure}
\section{Evaluation}
Positioning the Travel Planner Agent described in Section 4 as the data recipient, this section elucidates the impact of providing multi-granular personal data—derived from cross-service user browsing history—on the trade-off between personalization performance and privacy protection. Furthermore, from the perspective of service practicality, we quantitatively evaluate practical costs, specifically focusing on latency and token consumption.
\subsection{Dataset construction}
The evaluation dataset utilized web browsing histories, ranging from 50 to 110 items per persona, generated by 20 fictional personas designed to represent residents of Japan. The travel planning task in this study is premised on a scenario where personalized recommendations are derived from short-term context, specifically reflecting intensive information gathering performed by the user over the preceding 1 to 2 hours. To simulate this, we designed specific browsing tasks, each requiring approximately 30 minutes to complete (involving roughly 25 history items), and assigned three such tasks to each persona. Personas were defined by demographics including 10 distinct age groups (18 to 60), gender (male, female), residential addresses across Japan, and specific areas of interest. These areas of interest were diversified according to the persona settings (e.g., K-POP, business news, golf, hot springs, camping) to ensure that the evaluation tasks were not biased toward specific domains. Within these persona settings, we defined a "Profile" consisting of five elements: age, date of birth, gender, address, and name. In the evaluation experiments, this serves as the user's static basic information, equivalent to account registration data typically found in online services. Data collection involved eight Japanese participants (male and female), each role-playing 2 to 3 personas. They executed prescribed tasks based on their assigned persona attributes (e.g., Persona: 18-year-old female living in Chiba, interested in K-POP; Task: "Search for TWICE dance videos"). To minimize deviation from actual user behavior, this role-playing-based data collection was conducted exclusively with residents of Japan, using tasks specifically designed to reflect the context of living in Japan.
\begin{table*}[t]
  \caption{Evaluation results of personalization performance by LLM-as-a-Judge according to the type of personal data provided. Scores are on a 5-point scale, with higher values indicating better performance. While the Browsing History, which entails the highest privacy risk but reflects detailed user behavior, recorded the highest scores, Categories (Level 3) demonstrated performance closely approaching it, achieving 97.2\% of the Browsing History (Long)'s average score across the three metrics, despite having a lower privacy risk.}
  \label{tab:table2}
  \centering
  \includegraphics[width=\textwidth]{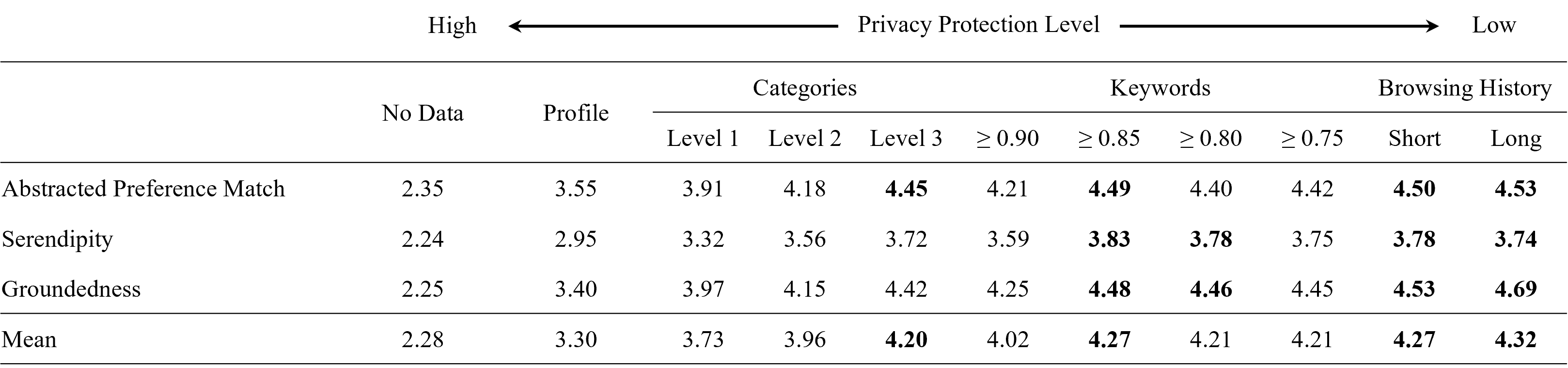}
\end{table*}
\subsection{Experimental Conditions}
To ensure the robustness of the evaluation and verify that the results are not dependent on specific constraints, we designed 5 distinct types of input Queries. Accordingly, we created Prompts requesting travel plans with varying durations (2 to 8 days) and corresponding budgets (25,000 to 250,000 JPY). Table 1 outlines the user context conditions. We defined 11 distinct user contexts based on the multi-granular privacy protection levels and the implementation patterns detailed in Section 4.2. A No Data condition was established as a baseline. Among the conditions where data is provided, the configurations range from a Profile-only setup to those that augment the Profile with data of varying granularities. For all conditions involving data provision, the Profile is included as a mandatory component. This design reflects the realistic assumption that users typically provide such basic information during account registration for standard services.

\subsection{Evaluation Method}
\subsubsection{Personalization Evaluation}
Evaluating personalization performance in travel planning is inherently challenging due to the difficulty of defining unique ground truths, rendering traditional automated metrics (e.g., n-gram similarity) insufficient. To address this, we adopted LLM-as-a-Judge \cite{zheng_judging_2023} as a flexible metric to evaluate the quality of the generated outputs. 

The evaluation prompt comprises the user context, the evaluation criteria, and the generated travel plan. These inputs are fed into the judge LLM, which outputs a score on a 5-point scale along with a rationale for each criterion. For the user context provided to the judge, we utilized the persona's basic information combined with the Detailed Browsing History (Long). We selected this configuration because the judging LLM requires a context that depicts user preferences in sufficient detail, and the Detailed Browsing History (Long) represents the data closest to raw logs, reflecting user behavior most comprehensively. Gemini 2.5 Flash was employed as the judge LLM.

We defined three evaluation criteria, scored on a 5-point scale. These criteria are grounded in literature emphasizing the importance of proposals based on abstracted preferences \cite{ramos_transparent_2024, zhao_personalens_2025}, the value of novel recommendations \cite{kotkov_rethinking_2023, mcnee_being_2006}, and the necessity of validity in the proposal's rationale \cite{maes_mitigating_2025, sani_fire_2025}.
\begin{itemize}
\item Abstracted Preference Match: Does the proposal satisfy user preferences abstracted from the history?
\item Serendipity: Does the proposal offer novel recommendations not present in the history?
\item Groundedness: Is the rationale for the response valid?
\end{itemize}
To mitigate potential bias regarding the evaluation scale, we explicitly included definitions for each score level within the evaluation prompt. Additionally, by enforcing a strict output schema on the Travel Planner Agent, we controlled the output lengths to be relatively uniform, thereby mitigating the verbosity bias of the LLM-as-a-Judge. Based on this methodology, we utilize the LLM-as-a-Judge results to quantify personalization performance. By correlating these results with the privacy protection levels corresponding to the conditions in Table 1, we evaluate the trade-off between personalization performance and privacy protection.
\subsubsection{Practical Cost Measurement}
Practical costs are evaluated using three metrics: latency, input token consumption, and output token consumption. Latency is measured as the total elapsed time from the moment the Travel Planner Agent receives the budget and schedule parameters until the travel plan is fully generated, encompassing the collaboration among multiple internal agents. Regarding token consumption, we aggregate the total input and output tokens utilized by all agents involved in the generation process. We measure these costs under each experimental condition to evaluate the impact of the data granularity on operational overhead.
\subsection{Preliminary Study}
To validate the LLM-as-a-Judge approach, we compared Gemini 2.5 Flash against human evaluators (N=8) across three metrics: Abstracted Preference Match, Serendipity, and Groundedness. Focusing on validation with restricted conditions, we found inter-rater reliability for single humans to be moderate (ICC(2,1)=0.47/0.33/0.44 respectively), whereas the mean of eight raters was high (ICC(2,\textit{k})\allowbreak=0.88\allowbreak/0.80\allowbreak/0.86), justifying the aggregated score as a reference. Correlation between LLM and human averages was positive but moderate (r=0.53/0.45/0.49; $\rho$=0.51/0.41/0.46). This partial alignment likely stems from the LLM’s strict adherence to criteria versus humans' reliance on implicit context. However, since the LLM successfully captures the relative performance trends across conditions, we conclude it is a robust metric for our primary goal: analyzing the comparative trade-off.
\begin{figure*}[t]
  \centering
  \includegraphics[width=\linewidth]{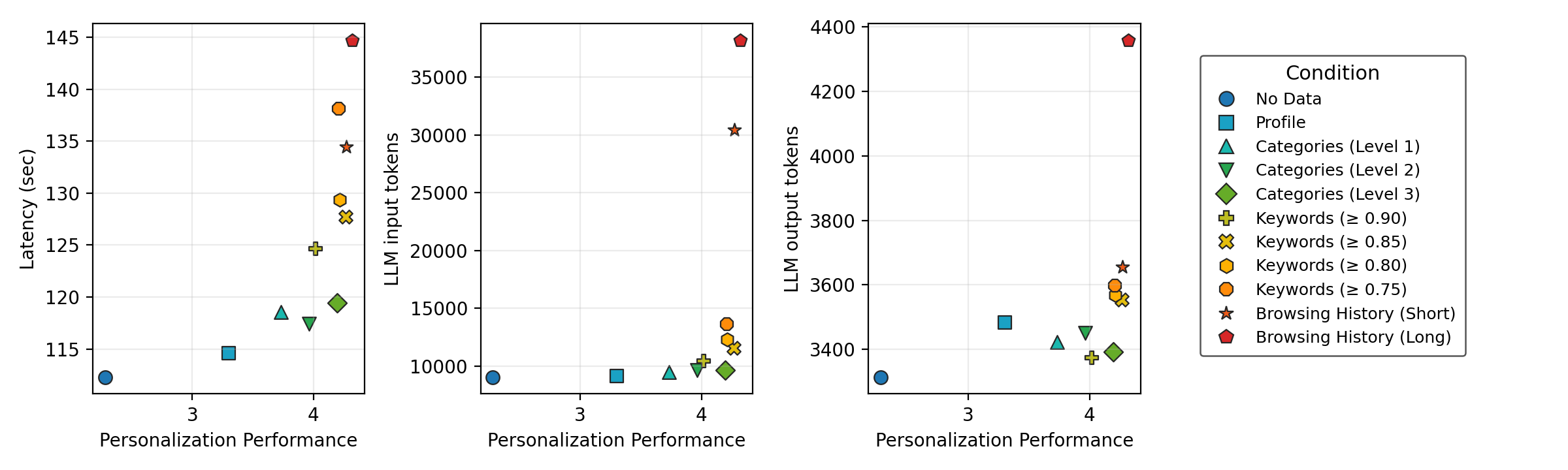}
  \caption{Latency and input/output token consumption during Travel Planner Agent inference under each personal data provision condition. The horizontal axis represents the mean score of the three-personalization metrics. All three scatter plots exhibit a similar trend, indicating that data with lower privacy protection levels tends to incur higher costs.}
  \Description{}
\end{figure*}
\section{Results}
\subsection{Personalization Performance}
Table 2 presents the results of the personalization performance evaluation conducted by the LLM-as-a-Judge across different personal data provision conditions. Overall, the results confirm the trade-off between personalization performance and privacy protection: conditions with lower privacy risks tend to yield lower personalization performance, whereas those with higher privacy risks achieve higher performance. However, analyzing specific data granularity patterns revealed findings that approach resolving this trade-off. Regarding Categories, while Level 1 and Level 2 underperformed compared to the Browsing History (Long) (which provides the most detailed context), Level 3 achieved an average score across the three metrics equivalent to 97.2\% of that of the Browsing History (Long), demonstrating performance closely rivaling the most detailed data. Notably, in terms of Abstracted Preference Match, Level 3 reached 98.2\% of the Browsing History (Long) score. Conversely, a distinct gap was observed in the Groundedness metric, confirming that detailed historical data remains superior for ensuring the validity of the rationale. For Keywords, the condition with a score threshold of $\geq 0.85$ yielded the best results across all three metrics among the keyword-based conditions. In keyword selection, there exists a trade-off between high information precision (selecting only high-scoring keywords) and information quantity (including lower-scoring keywords); our results suggest that the $\geq 0.85$ threshold represents the optimal balance point. Regarding Serendipity, keyword-based conditions recorded values exceeding those of the Browsing History, suggesting that for tasks such as travel planning, data at the granularity of keywords is sufficient—or potentially superior—for proposing novel content. The Browsing History conditions achieved the highest overall results among all categories. Furthermore, comparing Short and Long summaries, the Long version proved superior, indicating that the richness of information in the summaries contributes effectively to personalization performance.
\subsection{Practical Cost Measurement Results}
Figure 5 presents the input and output token consumption and latency observed during inference by the Travel Planner Agent under each personal data provision condition. The token consumption values represent the cumulative total, including all interactions among the agents involved in generating the travel plan. Both input and output token consumption demonstrated a clear trend: conditions with lower privacy risks resulted in lower consumption, while conditions with higher privacy risks resulted in higher consumption. Latency followed a similar trajectory, showing a strong correlation with token consumption.

While a significant disparity was observed between the No Data and Browsing History (Long) conditions, the results for Keywords and Categories exhibited some deviations, such as inversions relative to privacy risk levels or negligible differences depending on the specific pattern. Notably, Categories demonstrated relatively low operational costs, comparable to the No Data and Profile conditions. Considering the personalization performance results discussed in Section 6.1, these findings suggest a trade-off between personalization performance and practical cost. Section 7 will further discuss the effectiveness of the user-sovereign Puda architecture, including a detailed analysis of these relationships and considerations of conditions that approach resolving this trade-off.

\section{Discussion}
\subsection{Characteristics of Multi-Granular Data}
Based on the results from Section 6 regarding the multi-granular management of personal data collected across services, we identified a trade-off trend among personalization performance, privacy protection, and practical costs. Regarding the relationship between personalization performance and privacy protection, high performance was confirmed for Categories (Level 3), which deterministically restricts the output space. This Category (Level 3) also demonstrated superiority in terms of cost efficiency; both token consumption and latency were sufficiently low, approaching the levels observed in the No Data and Profile conditions. Consequently, Categories (Level 3) can be considered an effective data granularity that achieves a well-balanced compromise from a multifaceted perspective. These results suggest that content expressed by subsets of a predefined category list possesses the potential to realize personalization capabilities that closely approach those of personal data containing more specific details.

Conversely, when examining specific personalization metrics, the results suggest that Browsing History is effective for Groundedness, while Keywords are effective for Serendipity. The effectiveness of Browsing History for Groundedness is likely due to the specific behavioral history functioning effectively when explaining the rationale for a proposal. Regarding Serendipity, it is hypothesized that providing Keywords allowed the model to reference history on a word-level basis, thereby facilitating the grasp of the context and aiding in the distinction between known and novel items. From a multifaceted perspective, it appears that each data granularity offers distinct effects and benefits for personalized AI.
\subsection{User-Determined Data Granularity}
The efficacy of multi-granular personal data varies depending on the specific personalization task. This variation arises because the required granularity, data type, and the necessity of sensitive information differ inherently across tasks. In the travel planning task developed in this study, the objective was to propose destinations and POIs; thus, acquiring user preferences related to travel activities was sufficient. Consequently, word-based information such as Extracted Keywords derived from Browsing History or Predefined Category Subsets demonstrated high performance. Conversely, detailed behavioral history contained in the Browsing History is expected to manifest its full effectiveness in tasks requiring time-series information or in scenarios where the groundedness (rationale) of the proposal is paramount.

Regarding practical costs, their impact varies according to the frequency and context of task usage. High token consumption directly correlates with increased computational costs or external API usage fees. However, for tasks with low usage frequency, cost may not be a critical factor, potentially justifying a strategy that prioritizes performance improvement. Similarly, regarding latency, not all tasks require interactive, real-time responses; for comprehensive research or complex planning tasks, users may tolerate longer processing times in exchange for higher accuracy. Therefore, to realize true user sovereignty, users must be empowered to selectively utilize different data granularities based on their own intent and volition to achieve the appropriate level of personalization for their specific needs.
\subsection{Extensibility as a User-Sovereign Agent}
Fundamentally, Puda is designed to collect personal data across services and effectively provision it to personalized AI at privacy levels dictated by user instructions. While providing personalization functions as value-added services is pertinent in the current market, the scope of Puda extends far beyond this domain. Its core value lies in acting as the user's advocate. Consider, for instance, a scenario where a user schedules a dinner with a friend. In this context, respective Puda agents could interact to propose restaurants based on mutual preferences. It is precisely because Puda empowers users to determine privacy levels based on their own volition—rather than fully delegating these decisions to AI agents—that users can utilize autonomous personalized AI leveraging personal data with confidence. We anticipate that the range of autonomous tasks performed by AI agents will expand, leading to the increased utilization of personal data and the successive emergence of new value propositions centered on such data. In an era where personal data becomes the most critical asset, Puda stands as a user-sovereign agent, maximizing the utility of personal data while ensuring high security.

\section{Conclusion}
In this paper, we proposed Puda, a user-sovereign architecture that collects personal data across services and provisions it to personalized AI with privacy levels specified based on user judgment. This research is motivated by the urgent need for diverse, privacy-preserved user contexts for AI agents, addressing the current landscape where personal data is sequestered by dominant platform operators, thereby hindering cross-service utilization.

Specifically, Puda employs a browser-based approach to collect data across services and manages this data at multiple granularities. We evaluated the personalization performance using this data within a Travel Planner Agent task, employing an LLM-as-a-Judge methodology. The results confirmed an overall trade-off trend among personalization performance, privacy protection, and practical costs. However, we identified specific data granularities that approach resolving this trade-off. Notably, the Category (Level 3) condition achieved privacy protection via deterministic output space restriction and reasonable practical costs, while maintaining high personalization performance that closely rivaled (reaching 97.2\%) that of the Browsing History. Thus, we demonstrated the effectiveness of multi-granular personal data management by clarifying the distinct trade-off balances inherent to each data granularity. Furthermore, our discussion examined how these trade-offs vary depending on the task context, highlighting the critical importance of empowering users to determine these multi-granular privacy levels based on their own volition. This study bridges the gap between user-sovereign data management and the utilization of personal data for privacy-preserved AI agents, offering a blueprint for a user-sovereign, AI-native personal data management infrastructure.

\printbibliography
\end{document}